\documentclass[10pt,twocolumn,letterpaper]{article}

\usepackage[pagenumbers]{cvpr} 

\usepackage{color}

\usepackage{graphicx}
\usepackage{amsmath}
\usepackage{amssymb}
\usepackage{booktabs}
\usepackage{algorithm}
\usepackage{algorithmic}
\usepackage{multirow}
\usepackage{bbding}
\usepackage{threeparttable}

\usepackage[pagebackref,breaklinks,colorlinks]{hyperref}

\usepackage[capitalize]{cleveref}
\crefname{section}{Sec.}{Secs.}
\Crefname{section}{Section}{Sections}
\Crefname{table}{Table}{Tables}
\crefname{table}{Tab.}{Tabs.}

\def\cvprPaperID{8324} 
\def\confName{CVPR}
\def\confYear{2022}

\begin{document}

\title{Generalized Global Ranking-Aware Neural Architecture Ranker for Efficient Image Classifier Search}

\author{
Bicheng Guo\textsuperscript{1}, Tao Chen\textsuperscript{2}, Shibo He\textsuperscript{1}\thanks{Corresponding Author.}, Haoyu Liu\textsuperscript{3}, Lilin Xu\textsuperscript{1}, Peng Ye\textsuperscript{2}, Jiming Chen\textsuperscript{1}\\
\textsuperscript{1}Zhejiang University, \textsuperscript{2}Fudan University, \textsuperscript{3}Fuxi AI Lab, NetEase Games\\
{\tt\small \{guobc, s18he, cjm\}@zju.edu.cn}\\
{\tt\small \{eetchen, yepeng20\}@fudan.edu.cn}\\
{\tt\small liuhaoyu03@corp.netease.com}
}
\maketitle

\begin{abstract}
Neural Architecture Search (NAS) is a powerful tool for automating effective image processing DNN designing. The ranking has been advocated to design an efficient performance predictor for NAS. The previous contrastive method solves the ranking problem by comparing pairs of architectures and predicting their relative performance. However, it only focuses on the rankings between two involved architectures and neglects the overall quality distributions of the search space, which may suffer generalization issues. A predictor, namely Neural Architecture Ranker~(NAR) which concentrates on the global quality tier of specific architecture, is proposed to tackle such problems caused by the local perspective. The NAR explores the quality tiers of the search space globally and classifies each individual to the tier they belong to according to its global ranking. Thus, the predictor gains the knowledge of the performance distributions of the search space which helps to generalize its ranking ability to the datasets more easily. Meanwhile, the global quality distribution facilitates the search phase by directly sampling candidates according to the statistics of quality tiers, which is free of training a search algorithm, e.g., Reinforcement Learning~(RL) or Evolutionary Algorithm~(EA), thus it simplifies the NAS pipeline and saves the computational overheads. The proposed NAR achieves better performance than the state-of-the-art methods on two widely used datasets for NAS research. On the vast search space of NAS-Bench-101, the NAR easily finds the architecture with top 0.01\textperthousand~performance only by sampling. It also generalizes well to different image datasets of NAS-Bench-201, i.e., CIFAR-10, CIFAR-100, and ImageNet-16-120 by identifying the optimal architectures for each of them.
\end{abstract}

\begin{figure}[t]
  \centering
   \includegraphics[width=0.9\linewidth]{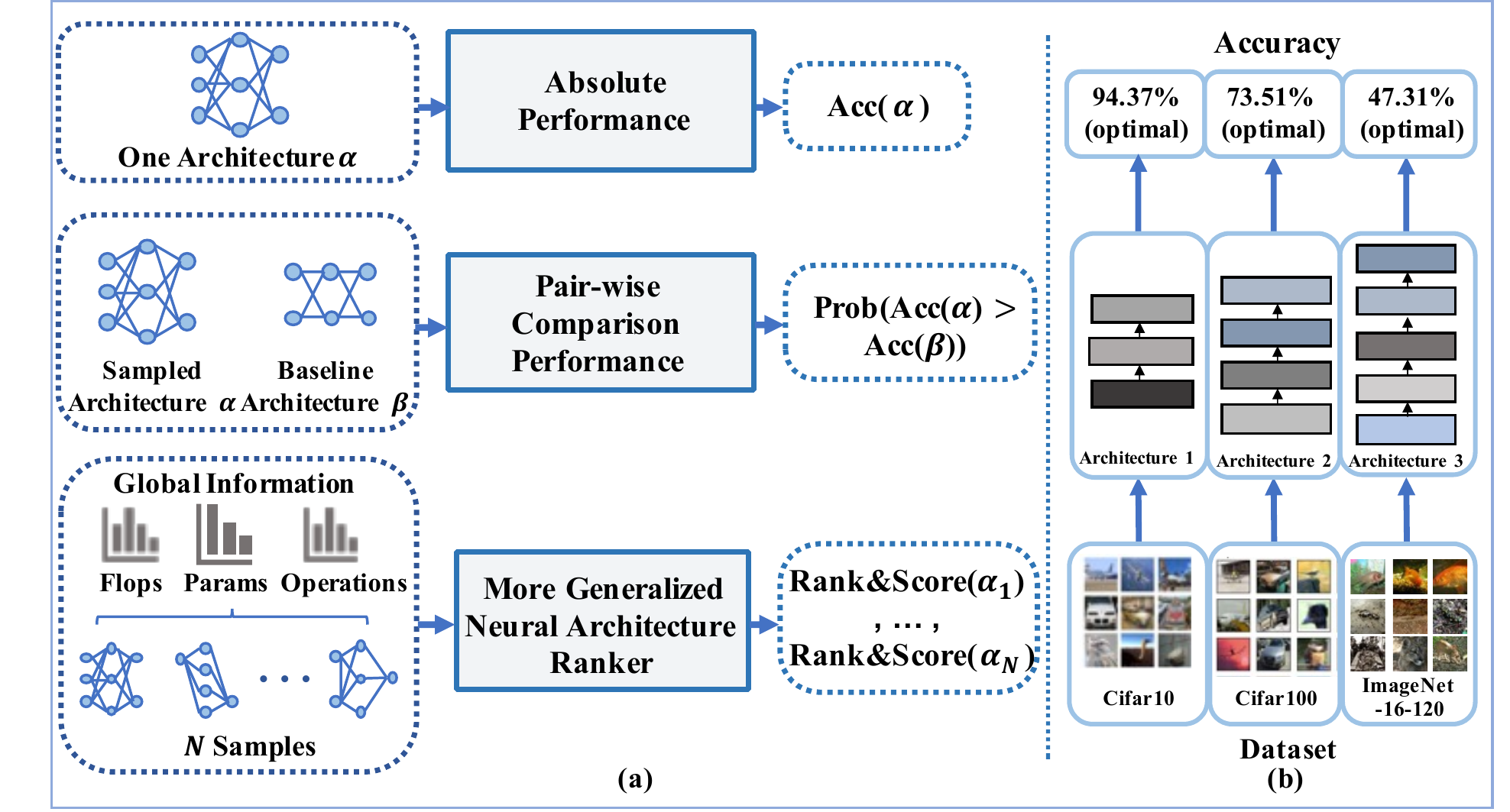}
   \caption{(a)~Comparison between previous performance predictors (top and middle) and ours (bottom). Different from predicting the absolute or pair-wise comparison performance, the NAR ranks the architectures with quality tier and scores them relative metric. (b)~ our NAR generalizes well to various image datasets by identifying the optimal architecture for each (CIFAR-10, CIFAR-100, and ImageNet-16-120).}
   \label{fig:illur}
\end{figure}

\section{Introduction}
\label{sec:intro}

Deep neural networks (DNNs) have received great attention in recent years. Many DNNs that are artificially designed by researchers have been applied to many scenarios, such as image classification~\cite{simonyan_2015_vgg,He_2016_resnet}, object detection~\cite{re_2015_faster, Redmon_2016_yolov1}, semantic segmentation~\cite{Long_2015_fcn, chen_2015_deeplabv1} and other real-world applications~\cite{Hasan_2021_elephant, Zhang_2021_monoflex, zhang_2021_fairmot}. Though these artificially designed DNNs have been proved to be powerful, designing them requires rich human expertise and is labor-intensive. Furthermore, dedicated knowledge is needed in the network architecture design for many specific target domains. 

Neural architecture search (NAS) offers a powerful tool for automating effective DNN designing for specific objectives. Previous studies directly apply different search and optimization methods, including Reinforcement Learning (RL)~\cite{Zoph_2017_rlnas,pham_2018_enas}, Evolutionary Algorithm (EA)~\cite{liu_2018_hierarchical, Real_2019_amoebanet}, and differentiable methods~\cite{liu_2018_darts, cai_2018_proxylessnas}, to find candidates in the search space. In order to reduce the prohibitive cost in evaluating a population of candidates, performance predictor is proposed to replace the evaluation metrics with the predicted performance of architectures~\cite{Luo_2018_nao, Cai_2020_ofa, Wen_2020_np}. 
However, these predictors try to approximate the absolute performance of architectures and suffer the ranking problem, i.e., the architectures with similar ground-truth performance have the incorrect predicted rankings due to the prediction bias, as shown in the top of Fig. \ref{fig:illur}~(a). As a result, the search algorithm could be misled to select the low-ranking architectures and yields the deteriorate results~\cite{Xu_2021_renas}.

The most recent contrastive method ~\cite{Chen_2021_ctnas} solves the ranking problem by comparing pairs of sampled architectures and calculating the probability that one architecture is better than the other, as shown in the middle of Fig. \ref{fig:illur}~(a). Even though it can achieve good ranking ability, such an approach may suffer from generalization issues. This is because it only focuses on the rankings between two involved architectures and neglects the overall quality distributions of the search space. As a result, this limits the predictor only to the architecture ranking instead of generalizing its ranking ability over various target datasets. To solve this, we propose to utilize the quality distributions of the search space where the architectures are classified into multiple quality tiers by their global rankings according to the ground-truth performance. In this way, we can train the predictor by learning the features of quality tiers and classifying each individual to the tier they belong to according to its global ranking. This classification enhanced ranking paradigm has also been explored previously~\cite{furnkranz2008multilabel}. The rational, \textit{class membership information is important for determining a quality ranking over a dataset}~\cite{Ji_ranking_kdd}, inspires us to first roughly classify the quality of the architecture and then score them. As a result, the predictor gains the global knowledge of the performance distributions of the search space which helps to generalize the ranking ability to various datasets more easily than previous local methods.

During the search phase, most of the previous studies either adopt reinforcement learning~\cite{Zoph_2017_rlnas, Tan_2019_mnas} or evolutionary algorithms~\cite{Guo_2020_spos, Lu_2021_nat}, which requires additional training cost and complicates the NAS pipeline. Interestingly, we can benefit from collecting the distributions of the top quality tiers and focus on the outperformed architectures by directly sampling with them in the search space. This makes our method free of training an RL controller or employing EA methods, thus saves the computational overheads.

In this work, we propose a Neural Architecture Ranker (NAR) to learn the global distributions of the architectures with various quality and identify each individual's quality level (tier) among the search space according to its performance (bottom of Fig. \ref{fig:illur}~(a)). 
Specifically, we first divide the search space into five quality tiers according to the performance distributions of the architectures. 
Then, each individual architectures is encoded to represent its structural and computational feature, and is matched with the embeddings of all tiers alternately to decide which tier it belongs to. 
In this way, we relax the performance prediction into quality classification problem. We also leverage the extracted feature to predict the relative scores of the sampled architectures. 
Consequently, the NAR is capable of ranking and scoring the candidates according to their global ranking among the search space, and generalizes well to various image datasets as shown in Fig. \ref{fig:illur}~(b). Furthermore, the distributions of the different quality tiers are collected to guide the sampling procedure in the search phase, which requires no additional computational overheads for searching and simplifies the NAS pipeline. The overall contribution is summarized as follows:
\begin{itemize}
    \item Different from locally comparing pairs of architectures and calculating relative probability, we propose a Neural Architecture Ranker (NAR) that ranks and scores the architectures by matching them with the representation of various quality tiers from a global perspective.
    \item We propose to collect the distributions with different quality tiers to guide the sampling in the search phase, which reduces cost compared with training an RL controller or employing EA method like before.
    \item We achieve state-of-the-art results on two widely used cell-based NAS datasets. On NAS-Bench-101, our NAR finds the architecture with top 0.01\textperthousand~performance among the vast search space of 423k individual architectures only by sampling, outperforming all other RL and EA methods. On NAS-Bench-201 with three different image datasets, i.e., CIFAR-10, CIFAR-100, and ImageNet-16-120, sharing the same search space, the NAR generalizes well to each of them by identifying the optimal architecture on a specific image dataset.
\end{itemize}

\section{Related Work}
\label{sec:related}
\subsection{Neural architecture search}

NAS offers to automate the design procedure of an efficient neural network given scenario constraints. It is often formulated as a constrained optimization problem:
\begin{equation}
\begin{aligned}
    &\min\limits_{\alpha \in \mathcal{A}} \mathcal{L}\left(W^{\ast}_{\alpha}; \mathcal{D}_{val}\right),\\
    \mathrm{s.t.} &W^{\ast}_{\alpha}=\arg\min\limits_{W_{\alpha}}\mathcal{L}\left(W_{\alpha};\mathcal{D}_{trn}\right), \\
    &\mathrm{cost}\left(\alpha\right)<\tau,
\end{aligned}
\label{eq:1stagegnas}
\end{equation}
where $W_{\alpha}$ are the weights of the architecture $\alpha$, $\mathcal{A}$ denotes the search space, $\mathcal{D}_{trn}$ and $\mathcal{D}_{val}$ mean the training and validation set respectively, $\mathcal{L}\left(\cdot\right)$ is the loss function, and $\mathrm{cost}\left(\alpha\right)$ denotes the computational cost with respect to $\alpha$, e.g., FLOPs, \#parameters\footnote{\#parameters denots the number of parameters.} or latency for different devices. Pioneering work applies RL~\cite{Zoph_2017_rlnas, pham_2018_enas,Tan_2019_mnas, Howard_2019_mbv3} and EA~\cite{real_2017_evo, liu_2018_hierarchical, Real_2019_amoebanet} to select $\alpha$ to evolve into the training and validation procedure which endures prohibitive cost.

In order to save the evaluation cost and simplify the optimization difficulties, two-stage NAS decouples the training and searching into two separate steps~\cite{Chu_2021_fair,Guo_2020_spos}. The first step is to jointly optimize all the candidates in one supernet:
\begin{equation}
    \min\limits_{W}\mathbb{E}_{\alpha\in\mathcal{A}}\left [ \mathcal{L}\left(W_{\alpha}; \mathcal{D}_{trn}\right)  \right ].
\label{eq:suernet}
\end{equation}
Then, RL or EA methods are utilized to select the subnet with best performance from the supernet given the constraints during the second search phase:
\begin{equation}
\begin{aligned}
    \left \{\alpha^{\ast}_i\right \} &=\mathop{\arg\min}\limits_{\alpha_i\in\mathbf{A}}\mathcal{L}\left(W^{\ast}_{\alpha_i}; \mathcal{D}_{val}\right)  \\
    &\mathrm{s.t.}~\mathrm{cost(\alpha_i)}<\tau_i, \forall~i,
\end{aligned}
\label{eq:search}
\end{equation}
where $W^{\ast}$ denotes the sharing weights inheriting from the trained supernet yielded in Eq. \ref{eq:suernet}.

Quite a lot of attention has been paid to improve the training quality and efficiency of the supernet in Eq. \ref{eq:suernet}, e.g., progressive shrinking~\cite{Cai_2020_ofa}, sandwich rule~\cite{Yu_2020_bignas}, and attentive sampling~\cite{Wang_2021_attentive}. During the search phase, all of existing methods either train an RL controller, or employ the EA to select the candidates with top quality. In this paper, we try to avoid such cost by collecting the distribution of outperformed architectures in the process of training the predictor and sampling according to them directly in the search phase.

Recently, vision transformers achieve state-of-the-art performance with powerful representation learning ability~\cite{Dosovitskiy_2021_vit}. AutoFormer~\cite{Chen_2021_ICCV_auto} is proposed to automate the non-trivial design process of choosing the suitable network depth, embedding dimension and head number by sharing the weights of transformer blocks. Moreover, the search space of the vision transformer is fine explored by evolving different dimensions, yielding superior performance~\cite{Chen_s3_2021}.

\begin{figure*}[t]
  \centering
   \includegraphics[width=\linewidth]{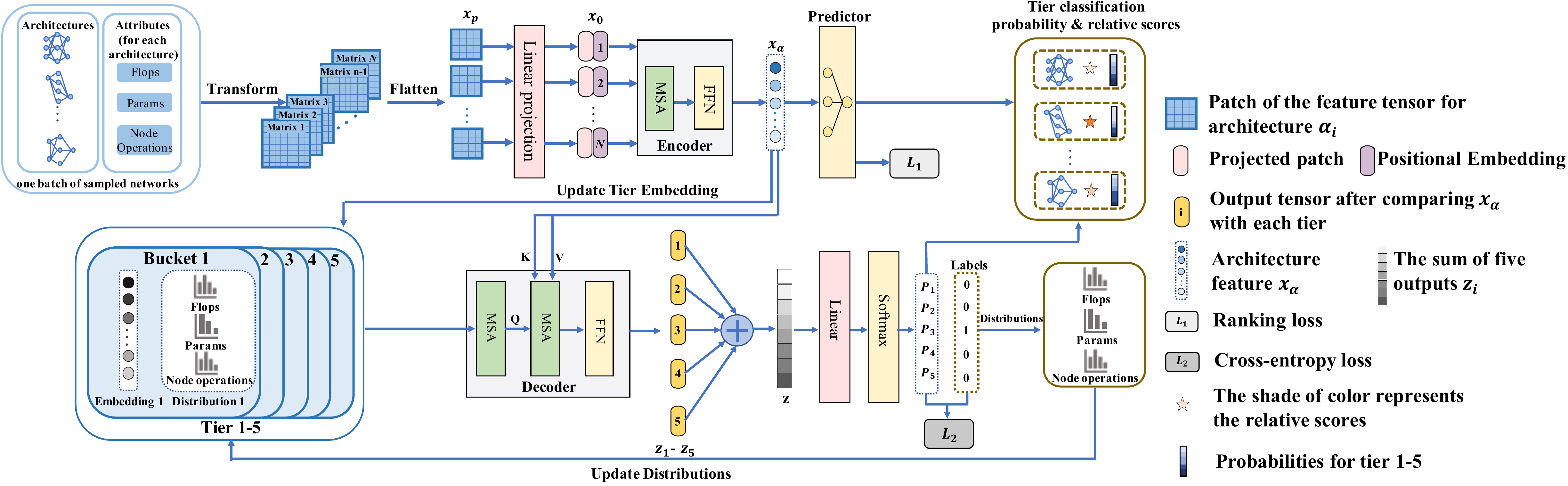}
   \caption{The overall training pipeline of the NAR. The NAR consists of encoder and decoder. The encoder extracts the feature of the sampled architecture and predicts its relative metric. The decoder matches the extracted feature with each of the tier embeddings and classifies it to the quality tier. The extracted features and batch distributions are updated to the corresponding tier according to the ground-truth performance of architectures.}
   \label{fig:ranker}
\end{figure*}
\subsection{Predictor and the Ranking problem}
To further boost the training and searching efficiency, the performance predictor is widely accepted both in one-stage and two-stage NAS~\cite{Liu_2018_pnas}. FBNetv3~\cite{Dai_2021_fbnetv3} applies a multi-layer perceptron to predict the accuracy and corresponding training recipe simultaneously. When training the supernet, predictor is utilized to facilitate the Pareto-aware sampling which improves the upper or lower performance bounds of the supernet~\cite{Wang_2021_attentive}. Predictor can also be used to generate the weights~\cite{zhang2018graph} or predict the latency of the network which pushes NAS to focus more on the hardware efficiency~\cite{Dudziak_2020_brpnas}.

Recently, ranking problem has attracted significant attention. For weight-sharing NAS, individual architectures with different parameters are actually shared with others in the supernet and this causes ranking disorder between standalone architectures and corresponding shared-weight networks. To tackle this, a set of landmark architectures with known standalone performance is selected to calculate a regulation term during the training of the weight-sharing supernet~\cite{Yu_2021_landmark}. Knowledge distillation is also a promising way to address this issue. During the supernet training, subnetworks with superior performance are selected as prioritized paths to transfer their knowledge to others, which flexibly eliminates the need for third-party models. Moreover, the overheads of training a searching algorithm is waived by directly selecting the best architectures from the prioritized paths~\cite{peng2020cream}.
For performance predictor, previous studies adopt the absolute performance as a metric to guide the following search procedure. However, the poor ranking correlation between ground-truth performance and evaluated metric of the architectures could deteriorate the search results, since the incorrect predicted rankings caused by the prediction bias will mislead the search algorithm to select the low-ranking architectures. In this case, the predictor based NAS algorithms focus on learning the relative ranking of the neural architectures instead of the absolute ones and can achieve the state-of-the-art results. ReNAS~\cite{Chen_2021_ctnas} learns the correct rankings between pairs of architectures by leveraging a ranking loss to punish the disordering predicted metric. CTNAS~\cite{Xu_2021_renas} directly compares two architectures and predicts the probability of one being better than the other. Both of the predictors limit to local pair-wise comparison, losing the overall picture of the entire search space. Besides, these comparators which learn a mapping relationship between pair-wise architectures and relative metric, heavily rely on the training data and may not generalize well to unseen architectures.

\section{Proposed Method}
\label{sec:problem}

We propose Neural Architecture Ranker~(NAR) to exploit the search space. We first relax the accuracy prediction into a quality classification problem. During the classification, the distributions of the representative parameters of the architectures, e.g., FLOPs, \#parameters and node operations, are then collected separately with respect to their quality. Finally, we utilize the distributions of the top tier to guide the sampling procedure, which is free of training an RL controller or employing EA in the search phase.

\subsection{Neural Architecture Ranker}
\label{subsec:nar}

In model design, we utilize the original Transformer~\cite{Vaswani_2017_trans} to implement the main procedures of \textit{matching and classifying}. Previous work only compares pair-wise architectures and learns a mapping relationship between the extracted features and metric, while we believe the more variety of architectures from the search space the predictor can handle, the more precise and confident the predictor examines the quality of the architecture. In this way, we first determine five different quality tiers $\mathcal{T}=\{T_1, T_2, T_3, T_4, T_5 \}$, according to the ground-truth performance of architectures.
For specific architecture $\alpha$, $\alpha \in T_1$ denotes that the architecture score is among the top 20\% rank. Then, we match each sampled architecture with the embedding which represents the population of the networks in each tier, and classify it into the corresponding tier. The training pipeline of the NAR is shown in Fig. \ref{fig:ranker}.

\textbf{Architecture encoding.} Similar to ReNAS, we encode each architecture of NAS-Bench-101 and NAS-Bench-201 datasets into feature tensor, $\mathcal{X} \in \mathbb{R}^{N \times P \times P}$, where $N$ denotes the number of the patches and $(P,P)$ is the patch resolution. We put the implementation of encoding in Appendix \ref{appd:archrepre}. We will release the detailed cell information datasets based on above datasets, including node FLOPs and \#parameters in each cell, as well as the feature tensor encoding codes to boost the NAS research one step further.

Once obtain the feature tensor, the $\mathcal{X}$ is reshaped into a sequence of flattened patches $\boldsymbol{x}_p \in \mathbb{R}^{N \times P^2}$. The $\boldsymbol{x}_p$ are then mapped to constant dimensions $D$ by a trainable linear projection which is similar to the ViT~\cite{Dosovitskiy_2021_vit}. We split the tensor into patches in a channel-wise way while ViT is along the image. Positional embeddings $\boldsymbol{E}_{pos}$ which helps keep the architecture macro skeleton information, are added to obtain the input embeddings $\boldsymbol{x}_{0}$ as
\begin{equation}
  \boldsymbol{x}_{0} = \boldsymbol{x}_p\boldsymbol{E} + \boldsymbol{E}_{pos},
  \label{eq:pos}
\end{equation}
where $\boldsymbol{E} \in \mathbb{R}^{P^2 \times D}$ denotes the weights of the linear projection, $\boldsymbol{E}_{pos} \in \mathbb{R}^{N \times D}$ adopts the original sine and cosine functions~\cite{Vaswani_2017_trans}.

\begin{figure*}[t]
  \centering
   \includegraphics[width=0.85\linewidth]{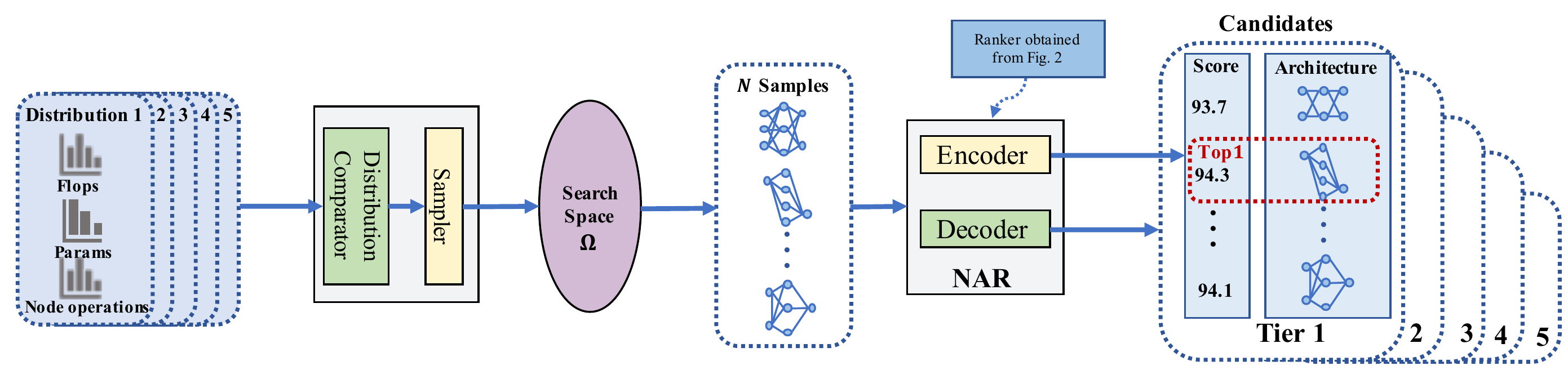}
   \caption{The sampling procedure in the search phase. First, the distribution of the top tier is used to compare with the ones in the last ranking tiers. Then, the full statistic or only its interval is utilized to sample in the search space. The sampled architectures are ranked and scored by the trained NAR. NAR selects the architecture with highest score in tier 1.}
   \label{fig:sampler}
\end{figure*}

\textbf{Supervised architecture feature extraction.} We utilize the sampled architecture features for three purposes: 1)~compare with the five different tier embeddings and decide which tire it belongs to; 2)~update the tier embeddings with the classified architecture; 3)~predict the relative metric of architecture to guide the selecting procedure in the search phase. The supervised architecture feature extracting is introduced due to its crucial role in the NAR.

We adopt the encoder to utilize self-attention for extracting needed feature. The encoder stacks $L=6$ identical layers and each layer consists of a Multiheaded Self-Attention (MSA) block and a fully connected Feed-Forward Network (FFN) block. LayerNorm (LN) is applied to each input of the block, and the block outputs are added with the value passed by the residual connection. We obtain the extracted architecture feature $\boldsymbol{x}_{\alpha} \in \mathbb{R}^{N \times D}$ by applying LN to the output feature of the last encoder layer (Eq. \ref{eq:enc_output}).
\begin{align}
    \boldsymbol{x}'_{l} &= {\rm MSA}({\rm LN}(\boldsymbol{x}_{l-1})) + \boldsymbol{x}_{l-1},\ &l = 1,\cdots, L, \label{eq:msa}\\
    \boldsymbol{x}_{l} &= {\rm FFN}({\rm LN}(\boldsymbol{x}'_{l})) + \boldsymbol{x}'_{l},\ &l = 1,\cdots, L,
    \label{eq:ffn} \\
    \boldsymbol{x}_{\alpha} &= {\rm LN}(\boldsymbol{x}_{L}). \label{eq:enc_output}
\end{align}

Inspired by the works~\cite{Chen_2021_ctnas,Yu_2021_landmark}, we employ two layers of linear projection with ReLU activation in-between on the feature $\boldsymbol{x}_{\alpha}$ to predict the relative metric $\hat{y}_\alpha$ of the architecture (neglect the bias),
\begin{equation}
    \hat{y}_\alpha = {\rm max}(0, \boldsymbol{x}_{\alpha}\boldsymbol{W}^1_r)\boldsymbol{W}^2_r
  \label{eq:val_acc}
\end{equation}
and adopt the ranking loss~\cite{Xu_2021_renas} for supervised training
\begin{equation}
    \mathcal{L}_1 = \sum_{m=1}^{k-1}\sum_{n=m+1}^{k}\psi((\hat{y}_{{\alpha}_m}-\hat{y}_{{\alpha}_n}) \ast {\rm sign}(y_{{\alpha}_m} - y_{{\alpha}_n})),
  \label{eq:ranking_loss}
\end{equation}
where $y_{{\alpha}_i}$ is the ground-truth accuracy of architecture $\alpha_i$, $k$ is the batch size, and $\psi(\epsilon) = {\rm log}(1+e^{-\epsilon})$ denotes the logistic loss function. With the supervised feature extraction, we expect that $\hat{y}_\alpha$ is able to prominently distinguish one architecture from others and obtain the correct rankings. This is critical for selecting the best network inside the top tier when sampling described in Section \ref{subsec:sampling}.

\textbf{Matching-based tier classification.} The decoder is utilized to extract tier embeddings which will be matched in turn with the sampled architecture feature and predict the probabilities of which quality tier it belongs to. The decoder stacks the same number of layers and each layer is inserted by one additional MSA block in addition to the two sub-layers of the encoder.

We initialize five tier embeddings, $\{\boldsymbol{e}_1, \boldsymbol{e}_2, \boldsymbol{e}_3, \boldsymbol{e}_4, \boldsymbol{e}_5\}$, where $\boldsymbol{e}_i \in \mathbb{R}^{N \times D}$. Each tier embedding $\boldsymbol{e}_i$ represents the architecture information of the corresponding tier $T_i$ and is added with the positional embeddings, yielding the input $\boldsymbol{z}^{0}_{i}$. The first MSA block of the decoder layer acts on $\boldsymbol{z}^{l}_{i}$ and outputs $\boldsymbol{q}^l_i$ as the query like the way in Eq. \ref{eq:msa}. We calculate the cross-attention by matching the sampled architecture feature $\boldsymbol{x}_{\alpha}$ with $\boldsymbol{q}^l_i$ in the second MSA block as
\begin{equation}
    \boldsymbol{z}'^{l}_{i} = {\rm MSA}({\rm LN}(\boldsymbol{q}^{l-1}_i), \boldsymbol{x}_{\alpha}) + \boldsymbol{q}^{l-1}_i,\ l = 1,\cdots, L.
    \label{eq:cross_att}
\end{equation}
Then $\boldsymbol{z}'^{l}_{i}$ goes through the FFN sub-layer of the decoder as Eq. \ref{eq:ffn} and in the last stack it yields the output $\boldsymbol{z}_i = {\rm LN}(\boldsymbol{z}^L_i)$. Notice that we apply above procedure five times to match the architecture with five different tier embeddings. All five outputs are summed up and we apply two layers of learnable linear transformation with ReLU in-between, as well as the softmax function to predict the probabilities of which tier the architecture belongs to:
\begin{equation}
    \boldsymbol{p}_{\alpha} = {\rm softmax}({\rm max}(0, \boldsymbol{z}\boldsymbol{W}^1_p)\boldsymbol{W}^2_p),
    \label{eq:prob}
\end{equation}
where $\boldsymbol{z}=\sum_{i=1}^{5}\boldsymbol{z}_i$ and $\boldsymbol{p}_{\alpha} \in \mathbb{R}^5$. We employ cross-entropy loss, denoted as $\mathcal{L}_2$, to jointly train the encoder and decoder of the NAR,
\begin{equation}
    \mathcal{L} = \mathcal{L}_2 + \lambda\mathcal{L}_1,
  \label{eq:total_loss}
\end{equation}
where $\lambda$ controls the importance between two different loss functions. In this way, the NAR manages to extract the feature of the sampled architecture, matching it with the embeddings of all five tiers and deciding its class.

\subsection{Tier representation and statistics}
\label{subsec:tier_update}

Once training one batch of $k$ sampled architectures is completed, their extracted features $\{\boldsymbol{x}_{{\alpha}_1}, \boldsymbol{x}_{{\alpha}_2}, \cdots, \boldsymbol{x}_{{\alpha}_k}\}$ obtained by Eq. \ref{eq:enc_output} are updated to corresponding tier embedding $\boldsymbol{e}_i$ according to their ground-truth tier labels (GT) by calculating the mean value of the features at the end of the iteration $t$:
\begin{equation}
    \boldsymbol{e}^t_i = \frac{\boldsymbol{e}^{t-1}_i\sum_{it=1}^{t-1}c^{it}_i + \sum{\boldsymbol{x}_{\alpha_i}}}{\sum^{t}_{it=1}c^{it}_i},
    \label{eq:update}
\end{equation}
where $\boldsymbol{x}_{\alpha_i} \in \{\boldsymbol{x}_{\alpha}\mid{\rm GT}({\alpha})=T_i\}$, $c^{it}_i$ denotes the counts of the architecture features belonging to $T_i$ at iteration $it$. This is similar to the case when we introduce a memory mechanism to store the features of sampled architectures and leverage them to match with the new ones. Notice that we update according to tier predictions during search phase since no labels are available during evaluating.

The distributions about FLOPs, \#parameters and node operations are also collected to guide the sampling procedure at the same time. For every batch, we discretize the interval of the FLOPs and \#parameters into $q$ constraints equally where step is calculated as
\begin{equation}
    \delta' = \frac{\tau_{\max}-\tau_{\min}}{q},
    \label{eq:step}
\end{equation}
where $\tau_{\max}$ and $\tau_{\min}$ are the maximum and minimum of the FLOPs or \#parameters in one batch. We round up the step $\delta = \lceil \delta' \rceil$ and then empirically approximate the distributions $\pi(\tau)$ as follows,
\begin{equation}
    \hat{\pi}(\tau_\kappa) = \frac{\#(\tau_\kappa=\tau_{\min} + \kappa\cdot\delta)}{k}, \ \kappa = 1, \cdots, q,
    \label{eq:distri}
\end{equation}
where $\tau_\kappa$ denotes the $\kappa_{\rm th}$ constraint on the interval, $\#(\tau_\kappa)$ denotes the number of the architecture located in constraint range $\left(\tau_{\kappa-1},\tau_{\kappa}\right]$ for FLOPs or \#parameters. As for every type of node operation, we just collect its total counts in each tier without discretization. Different from the offline way in AttentiveNAS~\cite{Wang_2021_attentive}, we train the NAR and simultaneously count the distributions for different quality tiers, $\{\hat{\pi}_{T_i}(\tau_\kappa)\mid i = 1, \cdots, 5; \kappa = 1, \cdots, q\}$. In this way, we have a thorough understanding about the hierarchical distribution of the search space for guiding the sampling process.

Above all, the Algo. \ref{alg:train} provides the overall training algorithm our Neural Architecture Ranker (NAR).

\begin{algorithm}[t]
    \caption{NAR Training Algorithm.}
    \label{alg:train}
    \begin{algorithmic}[1]
        \renewcommand{\algorithmicrequire}{\textbf{Input:}}
        \REQUIRE Search space $\Omega$; ${\rm encoder}$ and ${\rm decoder}$ of NAR.
        \STATE Randomly draw a set of architectures $\{\alpha^m\}$from $\Omega$.
        \STATE Obtain accuracy $y_{\alpha^{m}}$ by training $\alpha^m$
        \STATE Build $t$ batches of data, each batch $\{\alpha^m, y_{\alpha^{m}}\}^k_{m=1}$.
        \STATE Initialize five buckets $\{\phi_{T_1}, \cdots, \phi_{T_5}\}$ to hold corresponding tier's embedding and statistics, where $\phi_{T_i} = \{\boldsymbol{e}_{T_i}, \hat{\pi}_{T_i}^{\rm FLOPs}, \hat{\pi}_{T_i}^{\rm params}, \hat{\pi}_{T_i}^{\rm operations}\}$.
        \FOR{$iter \leftarrow 1:t$}
            \STATE Encode $\alpha^m$ into embedding $\boldsymbol{x}^{m}_0$.
            \STATE Build label $\boldsymbol{l}_i^m = \left[\mathbb{I}(\alpha^m \in T_1), \cdots, \mathbb{I}(\alpha^m \in T_5)\right]^{\rm T}$ by equally dividing the batch into five quality tiers according to $y_{\alpha^{m}}$.
            \STATE $\boldsymbol{x}_{\alpha^m}, \hat{y}_{\alpha^{m}} = {\rm encoder}(\boldsymbol{x}^{m}_0)$.
            \STATE Compute $\mathcal{L}_1 = \psi(\{\hat{y}_{\alpha^{m}}\}^k_{m=1}, \{y_{\alpha^{m}}\}^k_{m=1})$.
            \FOR{$i \leftarrow 1:5$}
                \STATE Get $i$th tier embedding $\boldsymbol{e}_i \in \phi_{T_i}$.
                \STATE $\boldsymbol{z}_i = {\rm decoder}(\boldsymbol{e}_i, \boldsymbol{x}_{\alpha^m})$.
            \ENDFOR
            \STATE Sum up $\boldsymbol{z}=\sum_{i=1}^{5}\boldsymbol{z}_i$ and predict probability $\boldsymbol{p}_{\alpha^m}$.
            \STATE Compute $\mathcal{L} = \mathcal{L}_2(\boldsymbol{p}_{\alpha^m}, \boldsymbol{l}_i^m) + \lambda\mathcal{L}_1$.
            \STATE Backward and update parameters of NAR.
            \STATE Update tier embeddings and statistics to all buckets.
        \ENDFOR
        \RETURN Buckets $\{\phi_{T_1}, \cdots, \phi_{T_5}\}$; trained NAR.
    \end{algorithmic}
\end{algorithm}

\subsection{Searching with tier statistics}
\label{subsec:sampling}

We utilize the trained NAR and the collected distributions to guide the search phase. Distributions are carefully selected first. Because, for some cases, one distribution depicting the top tier may be similar to those of low ranking tiers. Therefore, we apply Kullback-Leibler divergence to measure the difference between top and last ranking distributions as shown in Algo. \ref{alg:select_distri}. Specifically, once the divergence is less than the specified threshold, we discard the possibilities of distribution but sample architectures randomly on their interval. Noted that, this is not equivalent to discarding the entire distribution since we still sample on the interval where good candidates may locate in. Besides, we sample randomly on the interval in the same way when the population of the top tier is less than specific proportion of the sample size.

For every iteration, we sample $k$ subnets from the search space. Specifically we sample FLOPs and \#parameters as constrains with the selected distributions yielded in Algo. \ref{alg:select_distri} and reuse the constraints for every $m~(m<k)$ subnets. The implementation of the architectures sampling for NAS-Bench-101 and NAS-Bench-201 are detailed in Appendix \ref{subsec:imple}. After sampling the architecture, we build the subnet and reject those exceeding the constrains. In addition, we randomly sample certain proportion of subnets from the entire search space to increase the diversity.

Notice that the sampled architectures may not come from tier $T_1$ because each node operation is sampled independently and all these nodes are formed into the final architecture. In order to further narrow the searching phase to only the tier $T_1$ architectures, we first use the NAR to classify the sampled architectures and then score the ones from tier $T_1$. Specifically, giving a batch of $k$ samples, we leverage the trained NAR to rank and score the architectures, i.e., select top 5 architectures according to the predicted score in the classified tier $T_1$ as candidates at each iteration. Then, We train these candidates and obtain their evaluated accuracy. Thus, our sampling method fully exploits the knowledge of the training set to approximate the search space and achieves a top quality aware sampling procedure. The sampling procedure in the searching phase is shown in Fig. \ref{fig:sampler}.
\begin{algorithm}[t]
    \caption{Distribution Selection.}
    \label{alg:select_distri}
    \begin{algorithmic}[1]
    \renewcommand{\algorithmicrequire}{\textbf{Input:}}
    \REQUIRE Tier distributions $\{\hat{\pi}_{T_1}, \cdots, \hat{\pi}_{T_5}\}$; batch size $k$ and factor $\theta$ ; Kullback-Leibler divergence threshold $\zeta$; tier index $\beta~(\beta >1)$.
    \FOR{$i \leftarrow \beta:5$}
        \STATE Compute $d={\rm KL}(\hat{\pi}_{T_1}\parallel\hat{\pi}_{T_i})$.
        \IF{\#(samples) in $\hat{\pi}_{T_1} < \theta \cdot k$ \OR $d<\zeta$}
            \STATE Drop $\hat{\pi}_{T_1}$ possibilities but keep its interval $r_{T_1}$.
            \RETURN $r_{T_1}$.
        \ENDIF
    \ENDFOR
    \RETURN $\hat{\pi}_{T_1}$.
    \end{algorithmic}
\end{algorithm}

\section{Experiments}
\label{sec:exp}
In this section, we conduct extensive experiments to verify the effectiveness of the proposed NAR and sampling method on two widely accepted NAS search space, namely NAS-Bench-101~\cite{Ying_2019_nas101} and NAS-Bench-201~\cite{Dong_2020_nas201}. All our experiments are implemented on a single NVIDIA TITAN RTX GPU. The learning rate schedule and positional embeddings closely follow the settings from the Transformer~\cite{Vaswani_2017_trans}. We adopt AdamW~\cite{loshchilov_2018_decoupled} as the optimizer in our experiments. The implementation details are in Appendix \ref{subsec:imple}. The NAR codes and the detailed cell information datasets are available\footnote{https://github.com/AlbertiPot/nar.git}.

\begin{table*}[t]
    \caption{Searching results on the NAS-Bench-101 search space.}
    \centering
    \begin{tabular}{c|ccccc}
    \toprule
    \multirow{2}{*}{Methods} &
    \multirow{2}{*}{Average Accuracy (\%)} &
    \multirow{2}{*}{Best Accuracy (\%)} &
    \multirow{2}{*}{Best Rank (\textperthousand)} &
    \multicolumn{2}{c}{Cost (seconds)} \\
                                        &                   &                   &                   & train     & search    \\
\midrule
DARTS~\cite{liu_2018_darts}             & 92.21$\pm$0.61    & 93.02             & 13.47             & -         & -         \\
ENAS~\cite{pham_2018_enas}              & 91.83$\pm$0.42    & 92.54             & 22.88             & -         & -         \\
FairNAS~\cite{Chu_2021_fair}            & 91.10$\pm$1.84    & 93.55             & 0.77              & -         & -         \\
SPOS~\cite{Guo_2020_spos}               & 89.85$\pm$3.80    & 93.84             & 0.07              & -         & -         \\
FBNet~\cite{Wu_2019_fbnet}              & 92.29$\pm$1.25    & 93.98             & 0.05              & -         & -         \\
CTNAS\dag~\cite{Chen_2021_ctnas}     & 93.93$\pm$0.12    & 94.14    &          0.02              & 188.37    & 750.97    \\
ReNAS\dag~\cite{Xu_2021_renas}       & 93.93$\pm$0.09    & 94.02 & 0.09      & \textbf{73.68}    & -                     \\ 
\midrule
NAR (random)                            &  94.06$\pm$0.04  &   94.10   &    0.03    &     255.86     &      \textbf{53.51}     \\
NAR (statistics)                        &  \textbf{94.07$\pm$0.09}  &   \textbf{94.19}   &    \textbf{0.01}    &     292.61     &      189.94     \\
\bottomrule
    \end{tabular}
    \label{tab:nas101}
\begin{tablenotes}
    \footnotesize
    \item[1] ``Average Accurac'' denotes the average of the classification accuracy (\%) of the best architecture searched in each run on CIFAR-10 dataset. ``Best Accuracy'' and ``Best Rank'' denote the classification accuracy (\%) and thousandth rank of the best architecture searched in all runs. ``\dag'' denotes our implementation. All experiments are run 5 times.
\end{tablenotes}
\end{table*}

\subsection{Search results on NAS-Bench-101}
\label{subsec:nas101}

We verify the effectiveness of the proposed NAR method on the NAS-Bench-101 dataset. We randomly sample 1\% (4236) of the architectures and their averaged accuracy as our training set and another 1024 architectures as the validation set. Two types of sampling methods are tested: 1)~\textit{random}: randomly sample one batch of architectures from the entire search space in every iteration; 2)~\textit{statistics}: randomly sample a certain proportion of architectures and the rest of the batch are sampled with the collected distributions.

The proposed method is compared with the state-of-the-art NAS methods and main results are shown in Tab. \ref{tab:nas101}. All of the experiments are repeated 5 times with different random seeds. For fair comparison, we re-implement the recent state-of-the-art work~\cite{Xu_2021_renas, Chen_2021_ctnas} with the same random seeds and the training set size. As shown in Tab. \ref{tab:nas101}, the proposed NAR combined with \textit{random} or \textit{statistics} sampling method achieves new state-of-the-art performance on the measure of accuracy. With \textit{statistics} sampling, the proposed NAR framework outperforms other competitors on the average accuracy with relative low variance. Even more, it finds the individual architecture with top 0.01\% performance among the search space. The fact of achieving the superior performance is attributed to three reasons: 1)~the NAR is capable of well classifying the architectures of all tiers from the search space; 2)~the use of the collected distributions; 3) the adopted ranking loss is helpful to score the top architectures correctly. With \textit{random sampling}, the proposed NAR performs slightly worse than \textit{statistics} but still works better and more stable than the competitors. This demonstrates the superiority of the NAR for distinguishing top architectures given a random candidates. Without the collected distributions, it finds the individual architecture with top 0.03\textperthousand~performance among the search space, which is still competitive to others.

As for the search cost, the proposed sampling methods cost is lower compared to the CTNAS in which a LSTM is trained to generate architectures during searching. When sampling with \textit{random}, it achieves the lowest cost since it does not require rejection sampling. Different from encoding the architectures beforehand in ReNAS, our pipeline encodes architectures and trains the NAR simultaneously which costs more but is closer to the real application. ReNAS achieves the lowest training cost because it adopts LeNet-5 as the predictor while we utilize the advanced Transformer. To be noticed, the reported search cost is the system processing time rather than the computation cost. In real experiments, the proposed NAR actually requires no computation resource since it only performs sampling, while others cost GPU resources to train the RL or EA.
\begin{table*}[t]
    \caption{Searching results on the NAS-Bench-201 search space.}
    \centering
    \begin{tabular}{c|c|cccccc}
    \toprule
        \multirow{2}{*}{Methods} & \multicolumn{1}{c|}{Cost} &\multicolumn{2}{c}{CIFAR-10} & \multicolumn{2}{c}{CIFAR-100} & \multicolumn{2}{c}{ImageNet-16-120} \\
                      & \multicolumn{1}{c|}{(seconds)} & validation & test & validation & test & validation & test \\
        \midrule
        RSPS~\cite{li_2020_rsps}    &  7587  &   87.60$\pm$0.61  &   91.05$\pm$0.66   &  68.27$\pm$0.72   &   68.26$\pm$0.96   &  39.73$\pm$0.34   &   40.69$\pm$0.36   \\
        SETN~\cite{Dong_2019_setn}  &  31010  &  90.00$\pm$0.97   &   92.72$\pm$0.73   &   69.19$\pm$1.42  &   69.36$\pm$1.72   &  39.77$\pm$0.33   &   39.51$\pm$0.33   \\
        ENAS~\cite{pham_2018_enas}  &  13315  &  90.20$\pm$0.00   &  93.76$\pm$0.00    &   70.21$\pm$0.71  &   70.67$\pm$0.62   &  40.78$\pm$0.00   &   41.44$\pm$0.00   \\
        FairNAS~\cite{Chu_2021_fair}&  9845  &  90.07$\pm$0.57   &   93.23$\pm$0.18   &  70.94$\pm$0.94   &   71.00$\pm$1.46   &   41.90$\pm$1.00  &   42.19$\pm$0.31   \\
        ReNAS~\cite{Xu_2021_renas}   & 86.31 &90.90$\pm$0.31     &  93.99$\pm$0.25    &   71.96$\pm$0.99  &   72.12$\pm$0.79   &   45.85$\pm$0.47  &   45.97$\pm$0.49   \\
        GenNAS~\cite{li_2021_gennas} & 1080  &  -   &   94.18$\pm$0.10   &  -   &   72.56$\pm$0.74   &  -   &   45.59$\pm$0.54   \\
        DARTS-~\cite{chu2021dartsminus}&11520&   91.03$\pm$0.44  &   93.80$\pm$0.40   &   71.36$\pm$1.51  &   71.53$\pm$1.51   &   44.87$\pm$1.46  &   45.12$\pm$0.82   \\
        BOHB~\cite{falkner_2018_bohb}& \textbf{3.59}  &  91.17$\pm$0.27   &   93.94$\pm$0.28   &  72.04$\pm$0.93   &   72.00$\pm$0.86   &   45.55$\pm$0.79  &   45.70$\pm$0.86   \\
        \midrule
        NAR (average)              &  168.99 &  \textbf{91.44$\pm$0.10}   &   \textbf{94.33$\pm$0.05}   &  \textbf{72.54$\pm$0.44}   &   \textbf{72.89$\pm$0.37}   &  \textbf{46.16$\pm$0.37}   &   \textbf{46.66$\pm$0.23}   \\
        NAR (best)                 &  168.99 &       91.61       &        94.37       &       73.49       &       73.51        &        46.50      &     47.31            \\
        \midrule
        Optimal                    &    -    &        91.61      &        94.37       &       73.49       &       73.51        &        46.73      &     47.31            \\
        \bottomrule
    \end{tabular}
    \label{tab:nas201}
\begin{tablenotes}
    \footnotesize
    \item[1] ``validation'' and ``test'' denote the classification accuracy (\%) on the validation set and test set, respectively. ``average'' denotes the average of the classification accuracy (\%) of the best architecture searched in each run, ``best'' denotes the classification accuracy (\%) of the best architecture searched in all runs. ``Optimal'' denotes the highest accuracy for each set. In the search phase, only the interval of the collected distribution is adopted to sample architectures. The cost of the NAR framework reported includes the total cost of training and searching. All our experiments are run 10 times.
\end{tablenotes}
\end{table*}

\begin{table}
\caption{Comparisons of various number of quality tiers.}
\centering
\begin{tabular}{c|ccc}
\toprule
\multirow{2}{*}{\#tiers}  & top-$1$         & top-$5$ \\
         &  Avg. Acc.~(\%)     & Avg. Acc.~(\%)\\
\midrule
3       &     94.02$\pm$0.08~(94.10)            &     94.03$\pm$0.07~(94.10)   \\
5       &     94.02$\pm$0.12~(94.19)           &   94.07$\pm$0.09~(94.19)   \\
7       &     93.95$\pm$0.10~(94.11)            &  94.04$\pm$0.13~(94.19)   \\
\bottomrule
\end{tabular}
\label{tab:tiers}
\begin{tablenotes}
    \footnotesize
    \item[1] ``top-$k$'' denotes the $k$ architectures with the highest prediction scores. All experiments are run 5 times on the NAS-Bench-101 dataset. The best performance among the 5 runs is in the brackets.
\end{tablenotes}
\end{table}

\subsection{Search results on NAS-Bench-201}
\label{subsec:nas201}

We verify the generalization of the NAR on three image classification datasets from the NAS-Bench-201 dataset. We randomly sample 1000 architectures with their validation accuracy from the entire search space as our training set, and another 256 architectures as the validation set. The training and searching are the same with NAS-Bench-101 except for we sample randomly on the interval of the collected distribution since the search space is small. We compare the performance of the NAR with other NAS methods in Tab. \ref{tab:nas201}. The proposed NAR achieves new state-of-the-art average validation and test accuracy on all of three datasets with relative low variance. It demonstrates that our NAR is capable of finding the top quality architectures more stably compared to the competitors. Besides, adequate training samples which can offer useful distribution under acceptable evaluation cost are key to the good performance. Furthermore, the proposed method finds the optimal architecture which holds highest test classification accuracy among 3 datasets, proving the superiority and generalization ability of the propose NAR framework. The total cost consists of training at 126 seconds and search cost at 51 seconds, which are competitive.
\section{Ablation Study}

\subsection{Effect of the number of quality tiers}
\label{subsec:tiers}

\begin{figure}[t]
  \centering
   \includegraphics[width=0.85\linewidth]{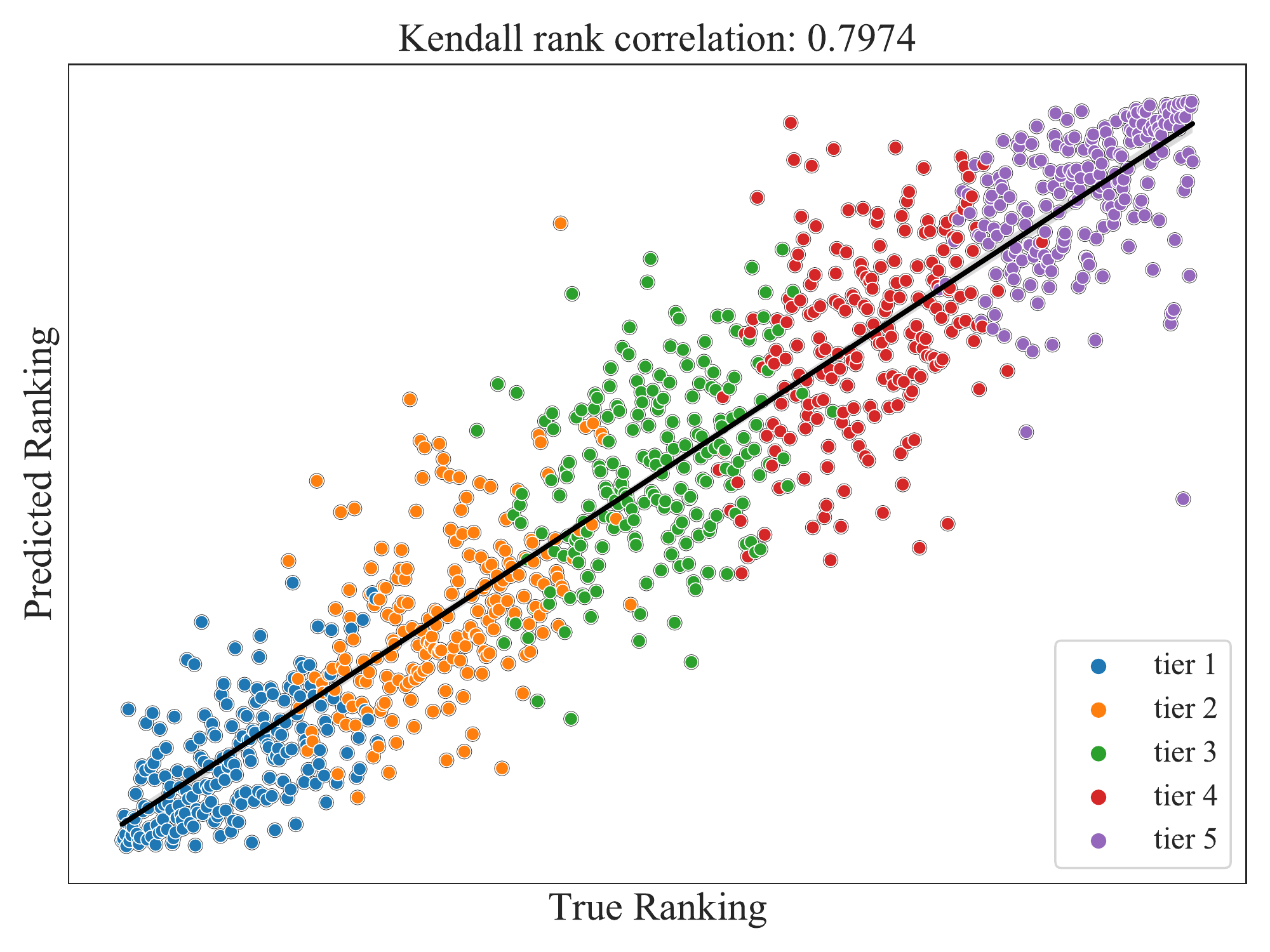}
   \caption{Rank correlation between the predicted and the actual ranking of the proposed NAR. 1024 architectures are randomly sampled from the NAS-Bench-101 dataset. Architectures are marked with their ground-truth teir. Kendall's $\tau=0.7974$ is obtained in one trial of 5 experiments.}
   \label{fig:ktau}
\end{figure}

We classify the quality distribution of the search space into 5 tiers. During the NAR training, it collects the statistics of the FLOPs and the \#parameters for each tier. These statistics will be used to sample candidates in the later search phase (Sec. \ref{subsec:sampling}). If the number of the tiers is set larger, the samples of each tier will be inadequate. If it is set smaller, each tier will cover multiple quality levels and could blur the useful classification information. We further visualize the tier distribution of a batch of 1024 architectures randomly sampled from the search space, and their classification results in Fig. \ref{fig:ktau}. It shows that when the number of tiers is set to 5, each tier has enough samples and they span uniformly in the search space, showing a clear quality distribution. We conduct ablation studies on NAS-Bench-101 to testify the above discussion. From the results shown in Tab. \ref{tab:tiers}, when the number of tiers is set to 5, the results are better than 3 and 7 in terms of average or the best performance.

\begin{table*}
\caption{Comparisons of the NAR with different losses.}
\centering
\begin{tabular}{c|ccc}
\toprule
\multirow{2}{*}{top-$k$}  & Cls. Loss & Ranking Loss & Cls.+Ranking Loss\\
                            &Avg. Acc.~(\%)&Avg. Acc.~(\%)&Avg. Acc.~(\%)    \\
\midrule
1       &       93.81$\pm$0.08(93.91)   &       94.00$\pm$0.08(94.11) &     94.02$\pm$0.12(94.19)  \\
3       &       93.86$\pm$0.04(93.91)   &       94.02$\pm$0.08(94.11) &     94.07$\pm$0.09(94.19)  \\
5       &       94.02$\pm$0.11(94.22)   &       94.02$\pm$0.08(94.11) &     94.07$\pm$0.09(94.19)   \\
\bottomrule
\end{tabular}
\label{tab:noloss}
\begin{tablenotes}
    \footnotesize
    \item[1] All experiments are run 5 times on the NAS-Bench-101 dataset. The performance of the best architecture among the 5 runs is in the brackets.
\end{tablenotes}
\end{table*}

\subsection{Effect of classification and ranking losses}

As discussed in Sec. \ref{sec:intro}, we apply classification and ranking loss simultaneously to learn a relative order of a batch of candidate architectures, as well as a global quality distribution of the search space. If we only adopt classification loss, the NAR roughly classifies the architectures into 5 tiers and chooses the candidate architectures according to the highest logits ($k$ =1, 3, 5) during the search phase. However, this logit obtained by classification loss merely indicates whether a sampled architecture belongs to a given tier and does not include their ranking information inside the tiers. So, we add a ranking loss to select the one with the top ranking from tier 1. When only adopting the ranking loss, we also consider the generalization problem that predictors may fail when they encounter architectures with large distinction~\cite{mehta2022nasbenchsuite}. To tackle this drawback, we adopt the classification loss to roughly pick up the candidates which is much more generalized than a pair-wised ranking loss. From the ablation results in Tab. \ref{tab:noloss}, we observe that, for ranking in the NAS, the combination of classification and ranking can improve the mean performance to some extent. Moreover, when only adopting ranking loss, the NAR is not able to find the superior architectures (94.11\% vs 94.19\%). This is possibly because the top-performing ones are so rare that the predictor with only pair-wise ranking loss cannot generalize well to identify them. More interestingly, when adopting the classification loss only, the predictor can find one with 94.22\% accuracy. This indicates that the predictor trained with classification loss can generalize well to identify those high-performed ones. However, when trained with the ranking loss (the right-most two columns ), the predictor cannot find the superior one.

\subsection{Effect of random samples}
\label{subsec:random_samples}

In order to investigate the effect of the proportion of the randomly sampled architectures, we conduct more experiments on the NAS-Bench-101 search space. As shown in Tab. \ref{tab:noisy}, when sampling entirely with the distribution collected during training ($p=0$), the NAR framework yields a deteriorate performance. However, with a certain proportion of randomly sampled architectures $p\in{0.3, 0.5, 0.7}$ added in or even all ($p=1$), the proposed method reaches a high mean accuracy. This may attribute to two points: 1)~the distribution is collected in each batch during the training, and distributions from all batches will be used by turns across all sampling iterations, so the samples of each distribution might not be adequate; 2)~The NAR is trained with different tiers of architectures while it can not compare and distinguish architectures from the same tier, so we propose to add certain proportion of randomly sampled architectures to the batch to improve the data diversity.
\begin{table}[t]
\caption{Comparisons of different proportions~($p$) of randomly sampled architectures.}
\centering
\begin{tabular}{c|cc}
\toprule
$p$   & Avg. Acc. (\%) & Best Acc. (\%) \\
\midrule
0.0 & 93.87$\pm$0.09        &    93.91       \\
0.3 & 93.99$\pm$0.10        &    94.17       \\
0.5 & \textbf{94.07$\pm$0.09}        &    \textbf{94.19}       \\
0.7 & 94.06$\pm$0.05        &    94.14       \\
1.0 & 94.06$\pm$0.04        &    94.10       \\
\bottomrule
\end{tabular}
\label{tab:noisy}
\begin{tablenotes}
    \footnotesize
    \item[1] All experiments are run 5 times on the NAS-Bench-101 dataset.
    \end{tablenotes}
\end{table}
\section{Conclusion}
In this work, we propose the Neural Architecture Ranker to rank and score the architectures for improving the searching efficiency of NAS. The NAR framework classifies the architectures into five different quality tiers and scores them with the relative metric. The tier distributions are collected to guide the sampling during the search phase which is free of excessive searching cost. Our methods outperforms previous NAS competitors on both the NAS-Bench-101 and NAS-Bench-201 datasets, stably finding the superior architectures from the search space with competitive cost. We will release two detailed cell information datasets to boost in-depth research into the micro structure in NAS field.

\section*{Acknowledgement}
This work was supported by the National Natural Science Foundation of China (No.~61790571, No.~61731004). We acknowledge the anonymous reviewers for their valuable comments. We thank Miao Guo for the helpful discussions and beautiful illustrations.

{\small
\bibliographystyle{ieee_fullname}
\bibliography{PaperForReview}
}

\appendix
\section{Architecture representation}
\label{appd:archrepre}

For NAS-Bench-101, each cell of the network contains at most 7 nodes, and there are total 9 cells. Nodes in each cell are denoted as operations and edges as connections. Following ReNAS~\cite{Xu_2021_renas}, cell connection is modeled by an adjacent matrix $\mathcal{A} \in \{0,1\}^{7 \times 7}$. When nodes are less than 7, we pad the missing rows and columns with 0. An operation type vector $\mathbf{o} \in \{1, \cdots,5\}^5$ is built to represent {\tt IN} node, $1 \times 1$ convolution, $3 \times 3$ convolution, $3 \times 3$ max-pooling and {\tt OUT} node, respectively. Besides, in order to represent the computational ability of the cell, the FLOPs and \#parameters of each node are calculated, and formed into FLOP vector $\mathbf{f} \in \mathbb{R}^7$ and \#parameters vector $\mathbf{p} \in \mathbb{R}^7$ for all nodes in each cell, where the missing nodes are padded with 0 when nodes of the cell are less than 7. Vectors $\mathbf{o}$ is first broadcast into matrices, and then element-wisely multiplied with the adjacent matrix $\mathcal{A}$ to form the operation matrix $\mathbf{O}$. Vectors $\mathbf{f}$ and $\mathbf{p}$ of each cell are transformed into the FLOPs matrix $\mathbf{F}$ and \#parameters matrix $\mathbf{P}$ in the same way. Finally, operation matrix $\mathbf{O}$ and matrices of all cells are concatenated into one feature tensor $\mathcal{X}$, whose size is (19, 7, 7).

For NAS-Bench-201, each node represents the sum of the feature maps and each edge as an operation. Every architecture has fixed 4 nodes and we build an adjacent matrix $\mathcal{A} \in \{0,1\}^{4 \times 4}$ without padding. The operation vector $\mathbf{o} \in \{0, \cdots,4\}^5$ represents zeroize, skip connect, $1 \times 1$ convolution, $3 \times 3$ convolution and $3 \times 3$ avg-pooling. The FLOPs vector $\mathbf{f} \in \mathbb{R}^4$ and the \#parameters vector $\mathbf{p} \in \mathbb{R}^4$ are obtained identically and broadcast into matrices. We concatenate those matrices according to the order of the cells and obtain the final feature tensor $\mathcal{X}$ with size (31, 4, 4). Every two matrices, from the second matrix to the last, correspond to the FLOPs and \#parameters of nodes for each cell, except for the first matrix representing the operations.

\begin{table*}
\caption{Comparisons of various properties of sampling.}
\centering
\begin{tabular}{c|ccc}
\toprule
\multirow{2}{*}{top-$k$}  & FLOPs & \#Params & FLOPs\&\#Params\\
                            &Avg. Acc.~(\%)&Avg. Acc.~(\%)&Avg. Acc.~(\%)    \\
\midrule
1       &       93.93$\pm$0.11(94.11)   &       93.97$\pm$0.11(94.17) &     94.02$\pm$0.12(94.19)  \\
3       &       93.99$\pm$0.11(94.14)   &       94.01$\pm$0.09(94.17) &     94.07$\pm$0.09(94.19)  \\
5       &       94.00$\pm$0.11(94.14)  &       94.01$\pm$0.09(94.17) &     94.07$\pm$0.09(94.19)   \\
\bottomrule
\end{tabular}
\label{tab:sample}
\begin{tablenotes}
    \footnotesize
    \item[1] All experiments are run 5 times on the NAS-Bench-101 dataset. The performance of the best architecture among the 5 runs is in the brackets.
\end{tablenotes}
\end{table*}

\section{Implementation}
\label{subsec:imple}

\textbf{Datasets.} We evaluate the NAR from two perspectives: 
\begin{enumerate}
\item{Test the validity of the proposed NAR to find the outperformed architectures from large search space.}
\item{Test the generalization of the NAR over different image datasets. This can be validated by finding the most suitable architecture for specific dataset from the same search space.}
\end{enumerate}

For the first perspective, we adopt the NAS-Bench-101~\cite{Ying_2019_nas101} dataset, which is a cell-based dataset containing over 423k unique convolutional architectures. All of the architectures are trained on the CIFAR-10 for 3 times to obtain the validation and test accuracy.

For the second, we employ the NAS-Bench-201 dataset~\cite{Dong_2020_nas201}. It is also a cell-based dataset with 15625 unique convolutional architectures and corresponding training, validation and test accuracy trained over CIFAR-10, CIFAR-100 and ImageNet-16-120 datasets.

\noindent \textbf{Settings.} For NAS-Bench-101, the AdamW optimizer is set with $\beta_1 = 0.9$, $\beta_2 = 0.982$, weight decay term is $5\times10^{-4}$ and $\epsilon = 10^{-9}$. The batch size is set to 256 and the NAR is trained for 35 epochs with 50 iterations as warm-up. For NAS-Bench-201, the AdamW optimizer is set with $\beta_1 = 0.9$, $\beta_2 = 0.99$, weight decay term is $1\times10^{-2}$ and $\epsilon = 10^{-9}$. The batch size is set to 128 and the NAR is trained for 55 epochs with 30 iterations as warm-up.

For the sampling details, the sample size is the same as the training batch size and set random samples rate at 0.5 to balance between the stable average accuracy and the superior individual selection ability, the ablation is at Section \ref{subsec:random_samples}. The constraints are reused for every 25 sampling trails and we sample total 50 iterations. For variables in Algo. \ref{alg:select_distri}, thresholds $\zeta$ of Kullback-Leibler divergence for FLOPs and \#parameters are both 2.5, the batch factor $\theta = 0.1$ and tier index $\beta = 4$.

For total 7 nodes of the cell in NAS-Bench-101, traversing from the second node (the first node is {\tt IN} node) to the last {\tt OUT} node, we first randomly sample from previous nodes of one specific node to build the connection, then sample the operation type of the node according to the collected distribution. For the remaining 3 edges (maximum 9 edges in each cell), two nodes are randomly sampled to build the connection and we repeat the procedure for 3 times. For NAS-Bench-201, since each cell has 4 nodes with fixed connection (each node connects to all of its previous nodes), we only sample the operation type for the edges of all nodes.

\begin{table}[t]
\caption{Comparisons of building block for predictor.}
\centering
\begin{tabular}{c|ccc}
\toprule
top-$k$ & Avg. Acc. (\%) & Best Acc. (\%) & Building block\\
\midrule
1     &      93.70$\pm$0.27 & 93.99     &  \multirow{3}{*}{ConvNet}\\
3     &      93.83$\pm$0.19 & 93.99    &   \\
5     &      93.89$\pm$0.16   & 94.04  &   \\
\midrule
1    &      94.02$\pm$0.12    & 94.19   & \multirow{3}{*}{MSA}\\
3   &       94.07$\pm$0.09    & 94.19 & \\
5    &      94.07$\pm$0.09    & 94.19  &  \\
\bottomrule
\end{tabular}
\label{tab:building}
\begin{tablenotes}
    \footnotesize
    \item[1] ``top-$k$'' denotes the $k$ architectures with the highest prediction scores. All experiments are run 5 times on the NAS-Bench-101 dataset.
\end{tablenotes}
\end{table}

\section{Effect of building block of the NAR}
\label{subsec:backbone}

Different from previous methods which adopt convolution network~\cite{Xu_2021_renas} as the predictor, we adopt the MSA block for the NAR because we want the predictor handles patches data, which is exactly our method for architecture encoding as mentioned in Section \ref{subsec:nar}. Specifically, the neural architecture is encoded into patches. MSA figures out which cell is important in the architecture and how it affects other cells by calculating the multi-attention score along the direction of the stacked patches. However, when using the convolution network, it operates on sub-regions inside the patches. This is equivalent to obtaining the relationship between different operations among different cells, which do not have direct connections in real architectures. Besides, we add positional embeddings to each patch to preserve the architecture depth information while the convolution network will make it invalid. To showcase how well the NAR adapts to architecture encodings, we compare our NAR with a 3-layer convolutional network trained with the ranking loss~\cite{Xu_2021_renas} to predict the ranking of the architectures and then sample candidates directly from the entire search space. The results in Tab. \ref{tab:building} show that MSA is the proper building block.

\section{Effect of sampling constraints}
\label{subsec:constrains}

As illustrated in Sec. \ref{subsec:sampling}, we sample the candidate architectures from the search space with the collected distributions of FLOPs and \#parameters following \cite{Wang_2021_attentive}. Since we do not apply any of search algorithms in the search phase, the quality of the sampling plays an indispensable role for selecting the top ranking architecture. Thus, we further ablate the  properties of the sampling procedure, i.e., we sample 1) only with FLOPs, 2) only with \#parameters and 3) with both. From the results  shown in Tab. \ref{tab:sample}, sampling with both properties yields better performance.

\begin{table}[t]
\caption{Comparisons of different number of the selected architectures during the search phase.}
\centering
\begin{tabular}{c|ccc}
\toprule
top-$k$ & Avg. Acc. (\%) & Queries & Ranking loss\\
\midrule
1     &      94.02$\pm$0.12 & 50     &  \multirow{5}{*}{\Checkmark}\\
3     &      94.07$\pm$0.09 & 150    &   \\
5     &      94.07$\pm$0.09   & 250  &   \\
7     &      94.07$\pm$0.09    & 350 &   \\
10    &      94.07$\pm$0.09    & 500 &   \\
\midrule
1    &      93.81$\pm$0.08    & 50   & \multirow{3}{*}{\XSolidBrush}\\
3   & 93.86$\pm$0.04        & 150 & \\
5    &      94.02$\pm$0.11    & 250  &  \\
\bottomrule
\end{tabular}
\label{tab:topk}
\begin{tablenotes}
    \footnotesize
    \item[1] ``top-$k$'' denotes the $k$ architectures with the highest prediction scores. ``Queries'' denotes the total number of queries to the ground-truth test accuracy for 50 iterations. All experiments are run 5 times on the NAS-Bench-101 dataset.
\end{tablenotes}
\end{table}

\section{Evaluate cost during sampling}

During the search phase, we sample for 50 iterations and select the top 5 architectures to query their test accuracies in every iteration, which are total 250 architectures to evaluate. In other words, the less architectures selected, the more evaluate cost we save. This requires the NAR holds high ranking and classification ability to ensure that we can find the outperforming candidates under limited trails. We compare the results of different number of the selected architectures in Tab. \ref{tab:topk}. It shows that the NAR can still achieve higher performance even with only one architecture selected, which dramatically reduces the search cost. We further investigate the results of the NAR trained without ranking loss, the average accuracy degrades more when one architecture selected. This demonstrates the ranking loss is essential to improve the ranking ability.

\section{The number of training samples}

\begin{figure}[t]
  \centering
  \includegraphics[width=0.85\linewidth]{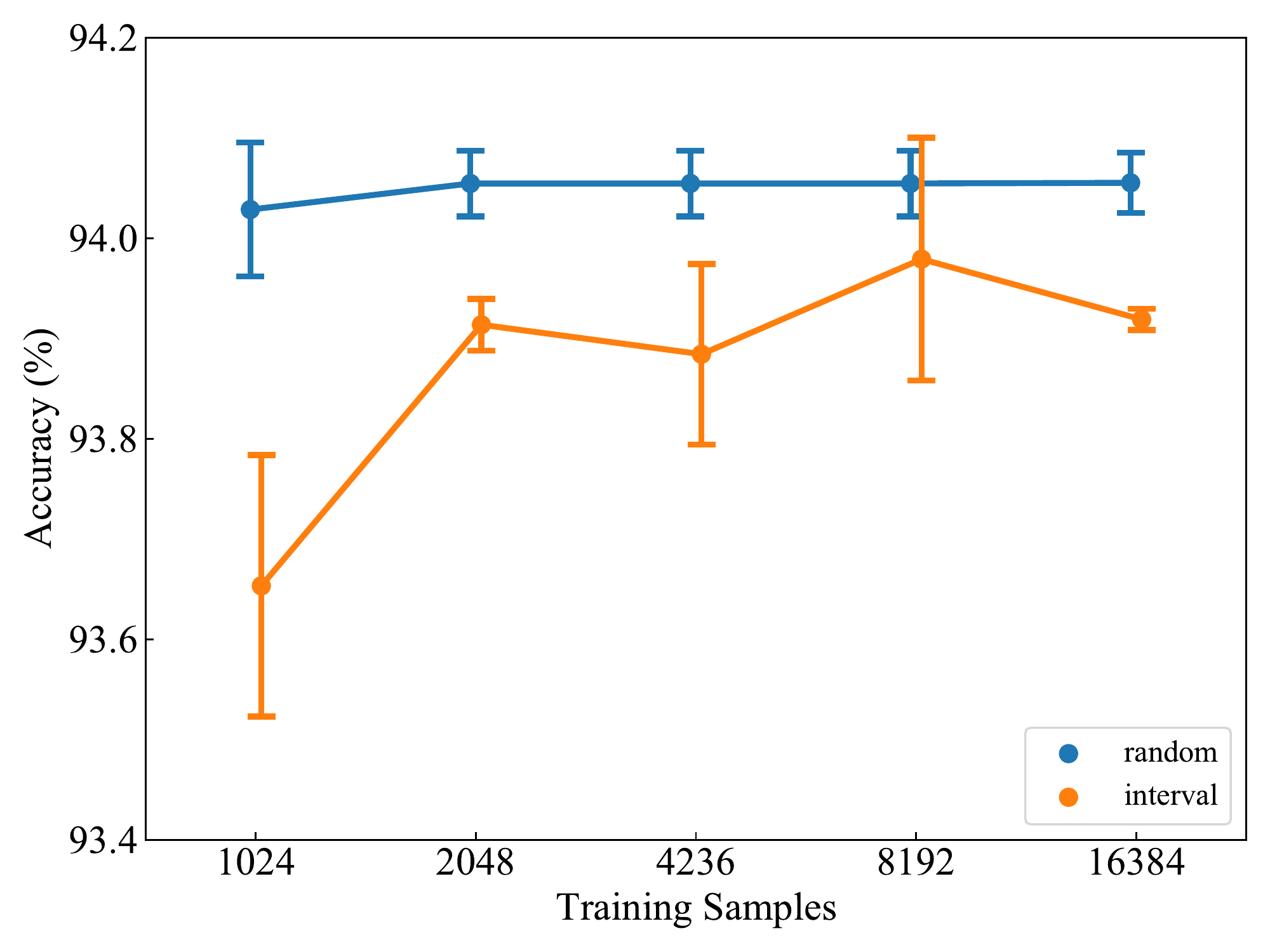}
  \caption{The effect of different number of training samples.
  ``random" denotes that architectures are randomly sampled from the search space; ``interval" denotes that architectures are randomly sampled on the interval of the distribution. All experiments are run 5 times on the NAS-Bench-101 dataset.}
\label{fig:trainingsize}
\end{figure}

The training samples are not only used to train the NAR, but also utilized to collect the distributions of all tiers. We further investigate the effect of the number of the training samples, i.e., train and collect on \{1024, 2048, 4236, 8192, 16384\} architectures randomly sampled from the NAS-Bench-101 dataset. Since the training size could affect the selection of the Kullback-Leibler divergence thresholds when sampling, we perform two kinds of random sampling during the search phase: 1)~randomly sample from the entire search space, denoted as \textit{random}; 2)~randomly sample a batch of architectures on the interval of the collected distribution in every iteration, denoted as \textit{interval}. As shown in Fig. \ref{fig:trainingsize}, when the samples size is 1024, the accuracy deteriorates significantly. Moreover, random sampling performs better and more stably than sampling on the interval. This requires that we add certain proportion of randomly sampled architectures as discussed in Section \ref{subsec:random_samples}.

\end{document}